\documentclass{article} 
\usepackage{nips13submit_e,times}
\usepackage{hyperref}
\usepackage{url}
\usepackage{graphicx}
\usepackage{amsmath}
\usepackage{amssymb}
\usepackage{amsthm}
\usepackage{mathabx}

\title{A Statistical Test for Joint Distributions Equivalence}

\author{
Francesco Solera \\
University of Modena, Italy \\
\texttt{francesco.solera@unimore.it} \\
\And Andrea Palazzi\\
University of Modena, Italy \\
\texttt{andrea.palazzi@unimore.it}
}

\newtheorem{theorem}{Theorem}
\newtheorem{definition}{Definition}

\nipsfinalcopy 

\begin{document}

\maketitle

\begin{abstract}
We provide a distribution-free test that can be used to determine whether any two joint distributions $p$ and $q$ are statistically different by inspection of a large enough set of samples.
Following recent efforts from Long~\emph{et al.}~\cite{long2016deep}, we rely on joint kernel distribution embedding to extend the \emph{kernel two-sample test} of Gretton~\emph{et al.}~\cite{gretton2012kernel} to the case of joint probability distributions. Our main result can be directly applied to verify if a dataset-shift has occurred between training and test distributions in a learning framework, without further assuming the shift has occurred only in the input, in the target or in the conditional distribution.
\end{abstract}

\section{Introduction}
\label{sec:preliminary}
Detecting when dataset shifts occur is a fundamental problem in learning, as one need to re-train its system to adapt to the new data before making wrong predictions. Strictly speaking, if $(X,Y)\sim p$ is the training data and $(X',Y')\sim q$ is the test data, a dataset shift occurs when the hypothesis that $(X,Y)$ and $(X',Y')$ are sampled from the same distribution wanes, that is $p\neq q$. The aim of this work is to provide a statistical test to determine whether such a shift has occurred given a set of samples from training and testing set.

To cope with the complexity of joint distribution, a lot of literature has emerged in recent years trying to approach easier versions of the problem, where the distributions were assumed to differ only by a factor. For example a covariate shift is met when, in the decomposition $p(x,y)=p(y|x)p(x)$, $p(y|x)=q(y|x)$ but $p(x)\neq q(x)$. Prior distribution shift, conditional shift and others can be defined in a similar way. For a good reference, the reader may want to consider Qui\~nonero-Candela~\emph{et al.}~\cite{quionero2009dataset} or Moreno-Torres~\emph{et al.}~\cite{moreno2012unifying}.

\section{Preliminaries}
\label{sec:pre_2}
As it often happens, such assumptions are too strong to hold in practice and do require an expertise about the data distribution at hand which cannot be given for granted. A recent work by Long~\emph{et al.}~\cite{long2016deep} has tried to tackle the same question, but without making restricting hypothesis on what was changing between training and test distributions. They developed the \emph{Joint Distribution Discrepancy} (JDD), a way of measuring distance between any two joint distributions -- regardless of everything else. They build on the \emph{Maximum Mean Discrepancy} (MMD) introduced in Gretton~\emph{et al.}~\cite{gretton2012kernel} by noticing that a joint distribution can be mapped into a tensor product feature space via kernel embedding.

The main idea behind MMD and JDD is to measure distance between distributions by comparing their embeddings in a Reproducing Kernel Hilbert Space (RKHS). RKHS $\mathcal{H}$ is a Hilbert space of functions $f: \Omega \mapsto \mathbb{R}$ equipped with inner products $\langle \cdot, \cdot \rangle_\mathcal{H}$ and norms $|| \cdot ||_\mathcal{H}$. In the context of this work, all elements $f\in\mathcal{H}$ of the space are probability distributions that can be evaluated by means of inner products $f(x) = \langle f, k(x,\cdot) \rangle_\mathcal{H}$ with $x\in\Omega$, thanks to the reproducing property. $k$ is a kernel function that takes care of the embedding by defining an implicit feature mapping $k(x,\cdot)=\phi(x)$, where $\phi:\Omega\mapsto\mathcal{H}$. As always, $k(x, x') = \langle \phi(x), \phi(x')\rangle_\mathcal{H}$ can be viewed as a measure of similarity between points $x, x' \in \Omega$. If a characteristic kernel is used, then the embedding is injective and can uniquely preserve all the information about a distribution~\cite{fukumizu2007kernel}.
According to the seminal work by Smola~\emph{et al.}~\cite{smola2007hilbert}, the kernel embedding of a distribution $p(x)$ in $\mathcal{H}$ is given by $\mathbb{E}_x [k(x, \cdot) ] = \mathbb{E}_x [\phi(x) ] = \int_\Omega \phi(x) \mathrm{d}P(x)$.

Having all the required tools in place, we can introduce the MMD and 
the JDD.
\begin{definition}[Maximum Mean Discrepancy (MMD)~\cite{gretton2012kernel}]
Let $\mathcal{F}\subset\mathcal{H}$ be the unit ball in a RKHS. If $x$ and $x'$ are samples from distributions $p$ and $q$ respectively, then the MMD is
\begin{equation}
\mathrm{MMD}(\mathcal{F},p,q) = \sup_{f\in\mathcal{F}}({\bf E}_x[f(x)]-{\bf E}_{x'}[f(x')])
\end{equation}
\end{definition}

\begin{definition}[Joint Distribution Discrepancy (JDD)~\cite{long2016deep}]
Let $\mathcal{F}\subset\mathcal{H}$ be the unit ball in a RKHS. If $(x,y)$ and $(x',y')$ are samples from joint distributions $p$ and $q$ respectively, then the JDD is
\begin{equation}
\label{eq:JDD_1}
\begin{split}
\mathrm{JDD}(\mathcal{F},p,q)   & = \sup_{f,g\in\mathcal{F}}({\bf E}_{x,y}[f(x)g(y)]-{\bf E}_{x',y'}[f(x')g(y')])\\
                                & = \|{\bf E}_{x,y}[\phi(x)\otimes\psi(y)] - {\bf E}_{x',y'}[\phi(x')\otimes\psi(y')]\|_{\mathcal{F}\otimes\mathcal{F}},
\end{split}
\end{equation}
where $\phi$ and $\psi$ are the mappings yielding to kernels $k_\phi$ and $k_\psi$, respectively.
\end{definition}

Note that, conversely to Long~\emph{et al.}~\cite{long2016deep}, we don't square the norm in Eq.~\eqref{eq:JDD_1}. A biased empirical estimation of JDD can be obtained by replacing the population expectation with the empirical expectation computed on samples $\{(x_1,y_1),(x_2,y_2),\dots,(x_m,y_m))\}\in X\times Y$ from $p$ and samples $\{(x'_1,y'_1),(x'_2,y'_2),\dots,(x'_n,y'_n))\}\in X'\times Y'$ from $q$:
\begin{equation}
\begin{split}
\mathrm{JDD}_b(\mathcal{F},X,Y,X',Y')   & = \sup_{f,g\in\mathcal{F}}\left(\frac{1}{m}\sum_{i=1}^mf(x_i)g(y_i)-\frac{1}{n}\sum_{i=1}^nf(x'_i)g(y'_i)\right)\\
& = \left\|\frac{1}{m}\sum_{i=1}^m\phi(x_i)\otimes\psi(y_i) - \frac{1}{n}\sum_{i=1}^n\phi(x_i')\otimes\psi(y_i')\right\|_{\mathcal{F}\otimes\mathcal{F}}\\
&=\left(\frac{1}{m^2}\sum_{i,j=1}^mk_\phi(x_i,x_j)k_\psi(y_i,y_j) + \frac{1}{n^2}\sum_{i,j=1}^nk_\phi(x_i',x_j')k_\psi(y_i',y_j')+\dots\right.\\
&\quad\left.-\frac{2}{mn}\sum_{i=1}^m\sum_{j=1}^nk_\phi(x_i,x_j')k_\psi(y_i,y_j')\right)^\frac{1}{2}
\end{split}
\end{equation}

Moreover, throughout the paper, we restrict ourselves to the case of bounded kernels, specifically $0\leq k(x_i,x_j)\leq K$, for all $i$ and $j$ and for all kernels.

\section{The test}
Under the null hypothesis that $p = q$, we would expect the JDD to be zero and the empirical JDD to be converging towards zero as more samples are acquired. The following theorem provides a bound on deviations of the empirical JDD from the ideal value of zero. These deviations may happen in practice, but if they are too large we will want to reject the null hypothesis.

\begin{theorem}
\label{theorem:test}
Let $p,q,X,X',Y,Y'$ be defined as in Sec.~\ref{sec:preliminary} and Sec.~\ref{sec:pre_2}. If the null hypothesis $p=q$ holds, and for simplicity $m=n$, we have
\begin{equation}
    \label{eq:test}
    \mathrm{JDD}_b(\mathcal{F},X,X',Y,Y') \leq \sqrt{\frac{8K^2}{m}(2-\log(1-\alpha))}
\end{equation}
with probability at least $\alpha$. As a consequence, the null hypothesis $p=q$ can be rejected with a significance level $\alpha$ if Eq.~\ref{eq:test} is not satisfied.
\end{theorem}

Interestingly, Type II errors probability decreases to zero at rate $\mathcal{O}(m^\frac{1}{2})$ -- preserving the same convergence properties found in the \emph{kernel two-sample test} of Gretton~\emph{et al.}~\cite{gretton2012kernel}. We warn the reader that this result was obtained by neglecting dependency between $X$ and $Y$. See Sec.~\ref{sec:exp} and following for a deeper discussion.

\section{Experiments}
\label{sec:exp}
\begin{figure}[h!]
\begin{tabular}{cc}
\hspace{1cm}\includegraphics[width=0.3\textwidth]{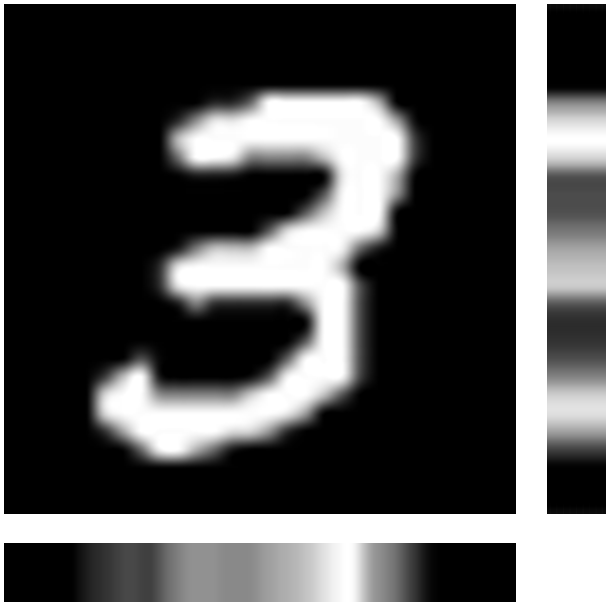} &\hspace{1.5cm}
\includegraphics[width=0.4\textwidth]{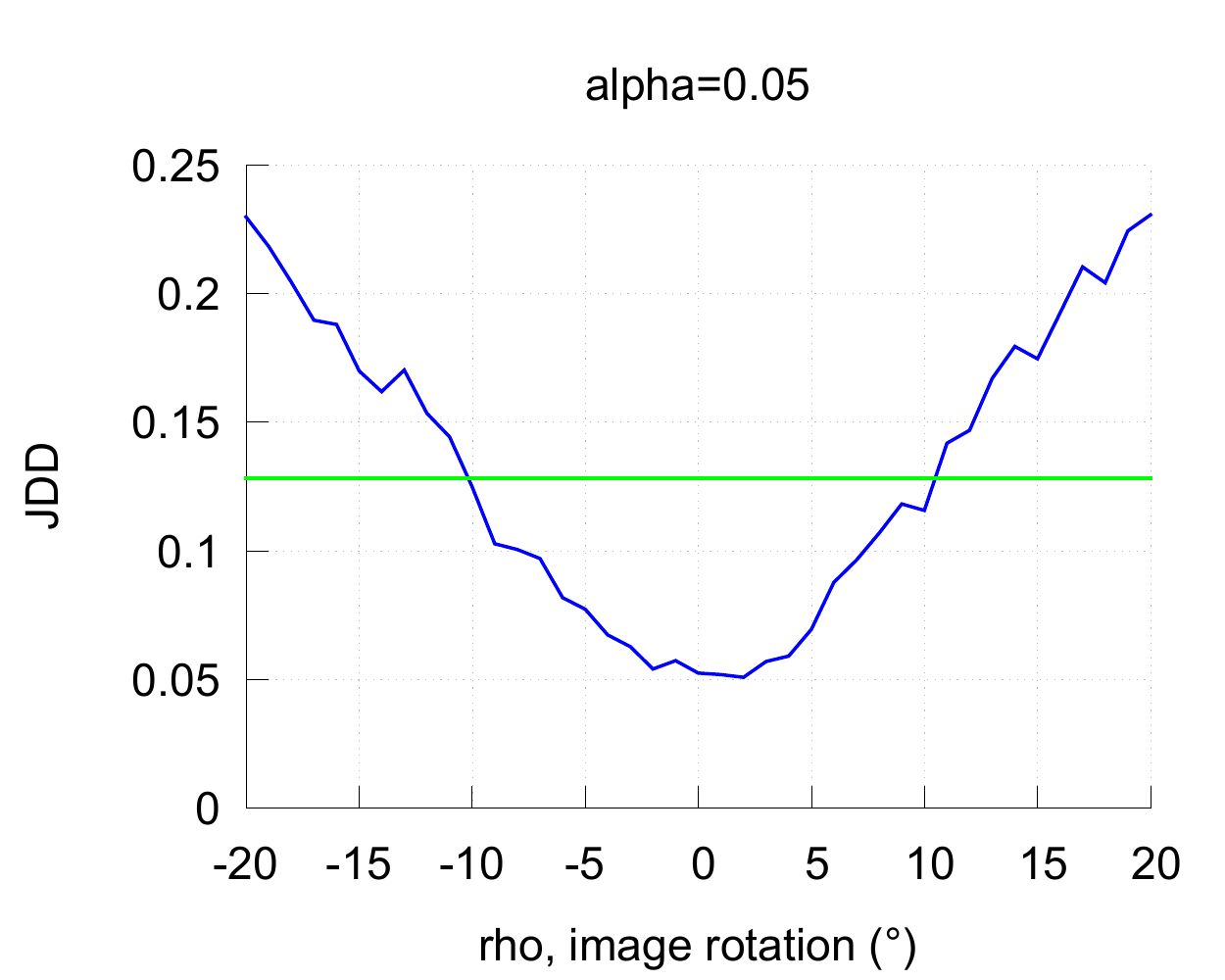}\\
\hspace{0.5cm}(a)&\hspace{2cm}(b)
\end{tabular}
\caption{In (a) we show an exemplar image drawn from the \texttt{MNIST} dataset. Observation $(x,y)$ are the projection histograms along both axis, \emph{i.e.} $x$ ($y$) is obtained by summing values across rows (columns). On the right, (b) depicts the behavior of the JDD measure when samples $(X',Y')$ are drawn from a different distribution w.r.t. $(X,Y)$, specifically the distribution of rotated images. The rotation is controlled by the $\rho$ parameter. The green line shows the critical value for rejecting the null hypothesis (acceptance region below).}
\label{fig:projection_hist_rot}
\end{figure}

To validate our proposal, we handcraft joint distribution starting from \texttt{MNIST} data as follows. We sample an image $i$ from a specific class and define the pair of observation $(x_i, y_i)$ as the vertical and horizontal projection histograms of the sampled image. Fig.~\ref{fig:projection_hist_rot}(a) depicts the process. The number of samples obtained in the described manner is defined by $m$ and they all belong to the same class. It is easy to see why the distribution is joint.
For all the experiments we employed an RBF kernel, which is known to be characteristic~\cite{gretton2006kernel}, \emph{i.e.} induces a one-to-one embedding. Formally, for $(x,y)$ and $(x',y')$ distributed according to $p$ or $q$ indistinctly, we have
\begin{equation}
k_\phi(x,x')= \exp\left(-\frac{\|x-x'\|^2}{\sigma_\phi^2}\right)\quad\quad\mathrm{and}\quad\quad k_\psi(y,y')= \exp\left(-\frac{\|y-y'\|^2}{\sigma_\psi^2}\right).
\end{equation}
The parameters $\sigma_\phi$ and $\sigma_\psi$ have been experimentally set to 0.25. Accordingly, both kernels are bounded by $K=1$.

In the first experiment we obtain $(X,Y)=\{(x_i, y_i)\}_{i}$ by sampling $m=1000$ images from the class of number $3$. Similarly, we collect $(X',Y')=\{(x_i', y_i')\}_{i}$ by applying a rotation $\rho$ to other $m$ sampled images from the same class. Of course, when $\rho = 0$, the two samples $(X,Y)$ and $(X',Y')$ come from the same distribution and the null hypothesis that $p=q$ should not be rejected. On the opposite, as $\rho$ increases in absolute value we expect to see JDD increase as well -- up to the point of exceeding the critical value defined in Eq.~\eqref{eq:test}. Fig.~\ref{fig:projection_hist_rot}(b) illustrates the behavior of JDD when $(X,Y)$ and $(X',Y')$ are sampled from increasingly different distributions.

\begin{figure}[h]
\centering
\begin{tabular}{cc}
\includegraphics[width=0.53\textwidth]{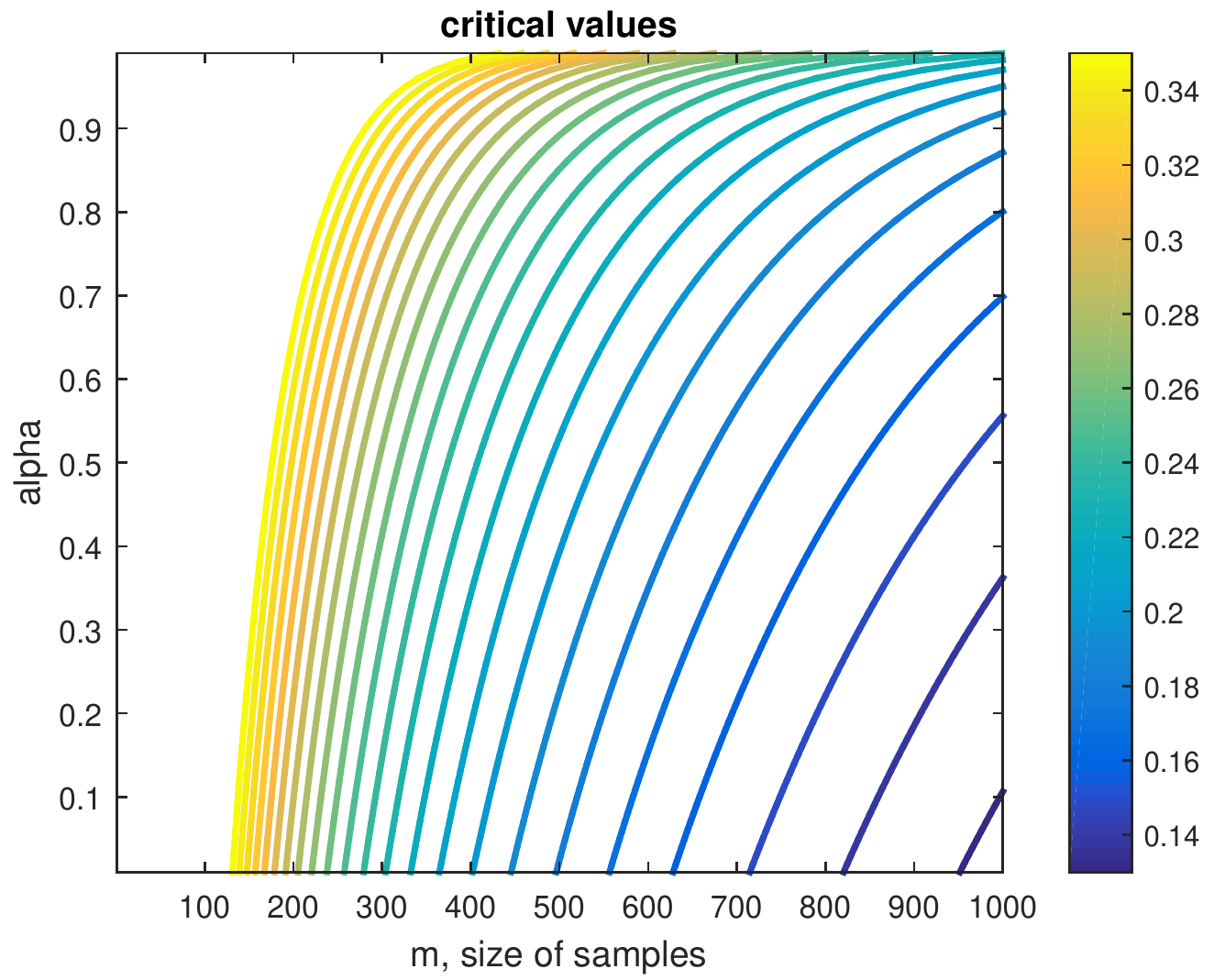} &
\includegraphics[width=0.45\textwidth]{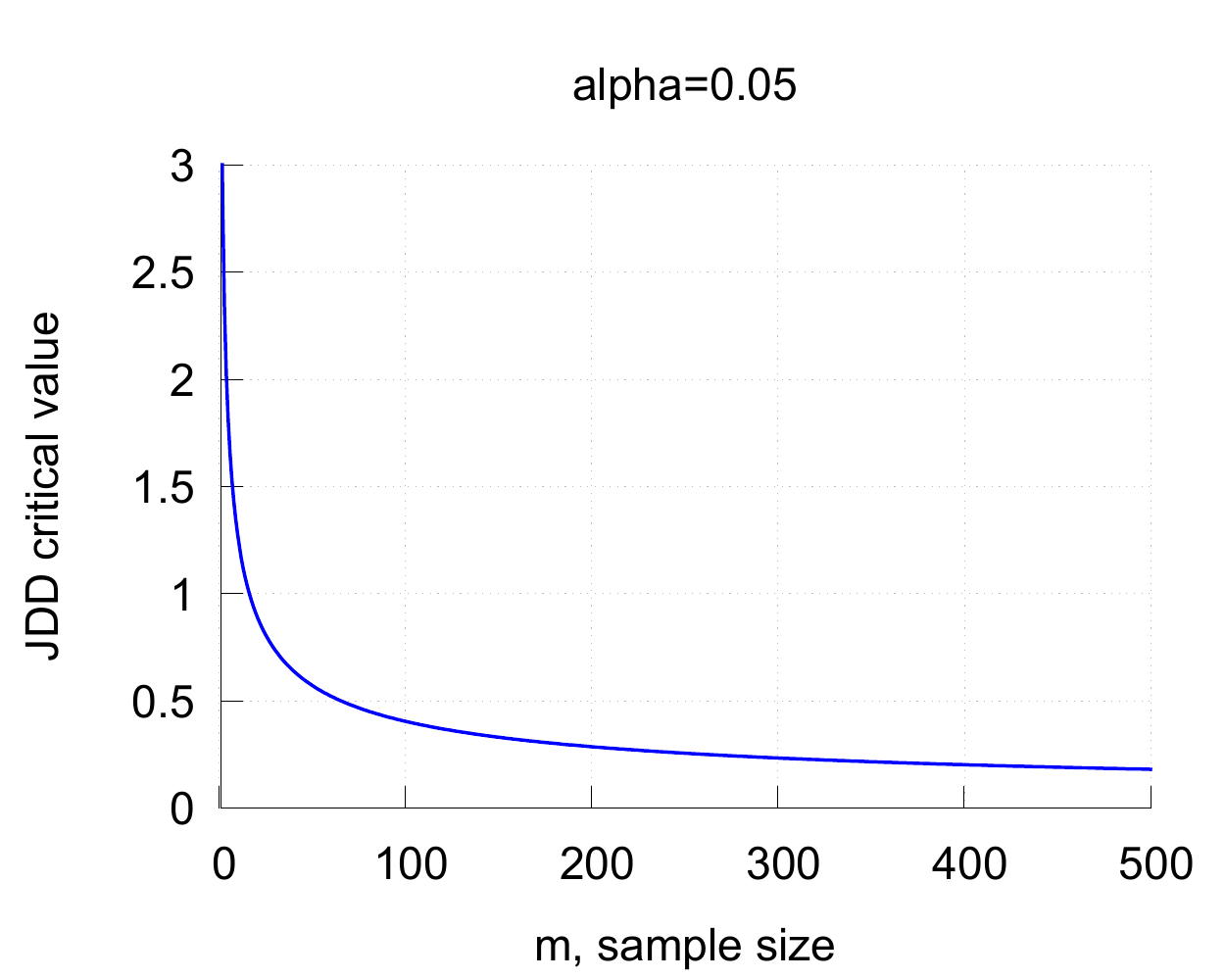} \\
(a)&(b)
\end{tabular}
\caption{On the left, (a) depicts the JDD critical value for the test of Eq.~\eqref{eq:test} when the significance level $\alpha$ and the sample size $m$ change. Cooler colors correspond to lower values of the JDD threshold. Not surprisingly, more conclusive and desirable tests can be obtained either by lowering $\alpha$ or by increasing $m$. Complementary, (b) shows the convergence rate of the test threshold at increasing size $m$ of sample, for a fix value of $\alpha=0.05$. It is worth noticing, that the elbow of the convergence curve is found around $m=50$.}
\label{fig:convergence}
\end{figure}

To deepen the analysis, in Fig.~\ref{fig:convergence} we study the behavior of the critical value by changing the significance level $\alpha$ and the sample size $m$.

\section{Limitations and conclusions}
\label{sec:lim}
The proof of Theorem~\ref{theorem:test} is based on the McDiarmid's inequality which is not defined for joint distributions. As a result, we considered all random variables of both distributions independent of each others, despite being clearly rarely the case. However, empirical (but preliminary) experiments show encouraging results, suggesting that the test could be safely applied to evaluate the equivalence of joint distributions under broad independence cases.

\newpage
\section{Appendices}

\subsection{Preliminaries to the proofs}
In order to prove our test, we first need to introduce McDiarmid's inequality and a modified version of Rademacher average with respect to the $m$-sample $(X,Y)$ obtained from a joint distribution.

\begin{theorem}[McDiarmid's inequality~\cite{mcdiarmid1989method}]
Let $f:\mathcal{X}^m\rightarrow\mathbb{R}$ be a function such that for all $i\in\mathbb{N}_m$, there exist $c_i<\infty$ for which
\begin{equation}
\sup_{X\in\mathcal{X}^m,\tilde{x}\in\mathcal{X}}|f(x_1, \dots,x_m)-f(x_1,\dots,x_{i-1},\tilde{x},x_{i+1},\dots,x_m)|\leq c_i.
\end{equation}
Then for all probability measures $p$ and every $\xi>0$,
\begin{equation}
\label{eq:mc_diarmid_result}
\mathrm{Pr}_X(f(X)-{\bf E}_X[f(X)] > \xi) < \mathrm{exp}\left(-\frac{2\xi^2}{\sum_{i=1}^mc_i^2}\right),
\end{equation}
where ${\bf E}_X$ denotes the expectation over the $m$ random variables $x_i\sim p$, and $\mathrm{Pr}_X$ denotes the probability over these $m$ variables.
\end{theorem}

\begin{definition}[Joint Rademacher average]
\label{def:rademacher_joint}
Let $\mathcal{F}$ be the unit ball in an RKHS on the domain $\mathcal{X}\times\mathcal{Y}$, with kernels bound between $0$ and $K$. Let $(X,Y)=\{(x_1,y_1),(x_2,y_2),\dots,(x_m,y_m)\}$ be an i.i.d. sample drawn according to probability measure $p$ on $\mathcal{X}\times\mathcal{Y}$, and let $\sigma_i$ be i.i.d. and taking values in $\{-1,+1\}$ with equal probability. We define the joint Rademacher average
\begin{equation}
\label{eq:rademacher_joint}
\hat{R}_m(\mathcal{F}, X, Y) = {\bf E}_\sigma\left[\sup_{f,g\in\mathcal{F}}\left|\frac{1}{m}\sum_{i=1}^m\sigma_i f(x_i)g(y_i)\right|\right].
\end{equation}
\end{definition}

\begin{theorem}[Bound on joint Rademacher average]
Let $\hat{R}_m(\mathcal{F}, X, Y)$ be the joint Rademacher average defined as in Def.~\ref{def:rademacher_joint}, then
\begin{equation}
\label{eq:rademacher_joint_bound}
\hat{R}_m(\mathcal{F}, X, Y) \leq K/m^{\frac{1}{2}}.
\end{equation}
\begin{proof}
The proof follows the main steps from Bartlett and Mendelson~\cite{bartlett2002rademacher}, lemma 22. Recall that $f(x_i)=\langle f,\phi(x_i) \rangle$ and $g(y_i)=\langle f,\psi(y_i) \rangle$, for all $x_i$ and $y_i$.
\begin{equation}
\begin{split}
\hat{R}_m(\mathcal{F}, X, Y) &= {\bf E}_\sigma\left[\sup_{f,g\in\mathcal{F}}\left|\frac{1}{m}\sum_{i=1}^m\sigma_i f(x_i)g(y_i)\right|\right]\\
&=\frac{1}{m}{\bf E}_\sigma\left[\left\|\sum_{i=1}^m\sigma_i\phi(x_i)\otimes\psi(y_i) \right\|\right]\\
&\leq\frac{1}{m}\left(\sum_{i,j=1}^m{\bf E}_{\sigma}\left[\sigma_i\sigma_jk_\phi(x_i,x_j)k_\psi(y_i,y_j)\right]\right)^\frac{1}{2}\\
&=\frac{1}{m}\left(\sum_{i}^m{\bf E}_{\sigma}\left[\sigma_i^2k_\phi(x_i,x_i)k_\psi(y_i,y_i)\right]\right)^\frac{1}{2}\\
&=\frac{1}{m}\left(\sum_{i}^mk_\phi(x_i,x_i)k_\psi(y_i,y_i)\right)^\frac{1}{2}\\
&\leq\frac{1}{m}(mK^2)^\frac{1}{2} = K/m^\frac{1}{2}
\end{split}
\end{equation}
\end{proof}
\end{theorem}

\subsection{Proof of Theorem 1}
We start by applying McDiarmid's inequality to $\mathrm{JDD}_b$ under the simplifying hypothesis that $m=n$, $$\mathrm{JDD}_b=\sup_{f,g\in\mathcal{F}}\left(\frac{1}{m}\sum_{i=1}^{m}f(x_i)g(y_i) - f(x_i')g(y_i')\right).$$
Without loss of generality, let us consider the variation of $\mathrm{JDD}_b$ with respect to any $x_i$. Since $\mathcal{F}$ is the unit ball in the Reproducing Kernel Hilbert Space we have
\begin{equation}
\label{eq:largest_var_f}
|f(x_i)| = |\langle{f, \phi(x_i)}\rangle| \le \|f\| \|\phi(x_i)\| \le 1 \times \sqrt{\langle{\phi(x_i),\phi(x_i)}\rangle} = \sqrt{k(x_i,x_i)} \le \sqrt{K}
\end{equation}
for all $f\in\mathcal{F}$ and for all $x_i$. Consequently, the largest variation to $\mathrm{JDD}_b$ is bounded by $2K/m$, as the bound in Eq.~\ref{eq:largest_var_f} also holds for all $g\in\mathcal{F}$ and for all $y_i$. Summing up squared maximum variations for all $x_i,y_i,x_i'$ and $y_i'$, the denominator in Eq.~\eqref{eq:mc_diarmid_result} becomes
\begin{equation}
4m\left(\frac{2K}{m}\right)^2 = \frac{16K^2}{m},
\end{equation}
yielding to
\begin{equation}
\label{eq:mcdiarmid_incomplete}
\mathrm{Pr}_{X,Y,X',Y'}(\mathrm{JDD}_b-{\bf E}_{X,Y,X',Y'}[\mathrm{JDD}_b] > \xi) < \mathrm{exp}\left(-\frac{m\xi^2}{8K^2}\right).
\end{equation}

To fully exploit McDiarmid's inequality, we also need to bound the expectation of $\mathrm{JDD}_b$. To this end, similarly to Gretton~\emph{et al.}~\cite{gretton2012kernel}, we exploit symmetrisation (Eq.~\eqref{eq:JDD_b_expectation}(d)) by means of a \emph{ghost sample}, \emph{i.e.} a set of observations whose sampling bias is removed through expectation (Eq.~\eqref{eq:JDD_b_expectation}(b)). In particular, let $(\widebar{X},\widebar{Y})$ and $(\widebar{X}',\widebar{Y}')$ be i.i.d. samples of size $m$ drawn independently of $({X},{Y})$ and $(X',Y')$ respectively, then

\begin{equation}
\label{eq:JDD_b_expectation}
\begin{split}
{\bf E}_{X,Y,X',Y'}[\mathrm{JDD}_b] &= {\bf E}_{X,Y,X',Y'}
\left[\sup_{f,g\in\mathcal{F}}\left(\frac{1}{m}\sum_{i=1}^{m}f(x_i)g(y_i) - \frac{1}{m}\sum_{i=1}^{m}f(x_i')g(y_i')\right)\right]\\
&\stackrel{(a)}{=}{\bf E}_{X,Y,X',Y'}
\left[\sup_{f,g\in\mathcal{F}}\left(\frac{1}{m}\sum_{i=1}^{m}f(x_i)g(y_i) - {\bf E}_{x,y}[fg] - \frac{1}{m}\sum_{i=1}^{m}f(x_i')g(y_i')+{\bf E}_{x',y'}[fg]\right)\right]\\
&\stackrel{(b)}{=}{\bf E}_{X,Y,X',Y'}
\left[\sup_{f,g\in\mathcal{F}}\left(\frac{1}{m}\sum_{i=1}^{m}f(x_i)g(y_i) - {\bf E}_{\widebar{X},\widebar{Y}}\left[\frac{1}{m}\sum_{i=1}^mf(\bar{x}_i)g(\bar{y}_i)\right] - \dots 
\right.\right.\\ &\left.\left.
\quad\quad\quad\quad\quad\quad\quad\quad\quad\quad\quad\quad
\frac{1}{m}\sum_{i=1}^{m}f(x_i')g(y_i')+{\bf E}_{\widebar{X}',\widebar{Y}'}\left[\frac{1}{m}\sum_{i=1}^mf(\bar{x}_i')g(\bar{y}_i')\right] \right)\right]\\
&\stackrel{(c)}{\leq}{\bf E}_{X,Y,X',Y',\widebar{X},\widebar{Y},\widebar{X}',\widebar{Y}'}
\left[\sup_{f,g\in\mathcal{F}}\left(\frac{1}{m}\sum_{i=1}^{m}f(x_i)g(y_i) - \frac{1}{m}\sum_{i=1}^mf(\bar{x}_i)g(\bar{y}_i) - \dots \right.\right.\\ &\left.\left.
\quad\quad\quad\quad\quad\quad\quad\quad\quad\quad\quad\quad
\frac{1}{m}\sum_{i=1}^{m}f(x_i')g(y_i')+\frac{1}{m}\sum_{i=1}^mf(\bar{x}_i')g(\bar{y}_i') \right)\right]\\
&\stackrel{(d)}{=}{\bf E}_{X,Y,X',Y',\widebar{X},\widebar{Y},\widebar{X}',\widebar{Y}',\sigma,\sigma'}
\left[\sup_{f,g\in\mathcal{F}}\left(\frac{1}{m}\sum_{i=1}^{m}\sigma_i(f(x_i)g(y_i) - f(\bar{x}_i)g(\bar{y}_i)) + \dots \right.\right.\\ &\left.\left.
\quad\quad\quad\quad\quad\quad\quad\quad\quad\quad\quad\quad
\frac{1}{m}\sum_{i=1}^{m}\sigma_i'(f(x_i')g(y_i')-f(\bar{x}_i')g(\bar{y}_i')) \right)\right]\\
&\leq{\bf E}_{X,Y,\widebar{X},\widebar{Y},\sigma}
\left[\sup_{f,g\in\mathcal{F}}\left|\frac{1}{m}\sum_{i=1}^{m}\sigma_i(f(x_i)g(y_i) - f(\bar{x}_i)g(\bar{y}_i))\right|\right] + \dots \\
&\quad\quad\quad\quad\quad\quad\quad\quad\quad\quad\quad\quad {\bf E}_{X',Y',\widebar{X}',\widebar{Y}',\sigma'}
\left[\sup_{f,g\in\mathcal{F}}\left|\frac{1}{m}\sum_{i=1}^{m}\sigma_i'(f(x_i')g(y_i') - f(\bar{x}_i')g(\bar{y}_i'))\right|\right]\\
&\leq2{\bf E}_{X,Y}[\hat{R}_m(\mathcal{F}, X, Y)]+2{\bf E}_{X',Y'}[\hat{R}_m(\mathcal{F}, X', Y')]\\
&\leq4K/m^\frac{1}{2}.
\end{split}
\end{equation}
In Eq.~\eqref{eq:JDD_b_expectation}, (b) adds a difference ${\bf E}_{x,y}[fg]-{\bf E}_{x',y'}[fg]$ that equals 0 since $p=q$ by the null hypothesis, and (c) employs Jensen's inequality.

\noindent By substituting the upper bound of Eq.~\eqref{eq:JDD_b_expectation} in Eq.~\eqref{eq:mcdiarmid_incomplete}, we obtain Theorem~\ref{theorem:test}.


\bibliographystyle{unsrt}
\bibliography{main}

\end{document}